\documentclass[
twocolumn,
]{ceurart}

\usepackage{listings}
\usepackage{subcaption}
\usepackage{graphicx}
\usepackage{hyperref}
\begin{document}

\copyrightyear{2023}
\copyrightclause{Copyright for this paper by its authors.
  Use permitted under Creative Commons License Attribution 4.0
  International (CC BY 4.0).}

\conference{SwissText 2023 submission}

\title{Voting Booklet Bias: Stance Detection in Swiss Federal Communication}

\author[1]{Eric Egli}[%
email=eric.egli@uzh.ch,
url=https://er.iceg.li,
]
\fnmark[1]
\address[1]{University of Zurich, Switzerland}

\author[1]{Noah Mamié}[%
email=noah.mamie@uzh.ch,
url=https://nmamie.github.io/,
]
\fnmark[1]

\author[1]{Mathias Müller}[%
email=mmueller@cl.uzh.ch,
]

\author[1]{Eyal Liron Dolev}[%
email=eyalliron.dolev@uzh.ch,
]

\fntext[1]{These authors contributed equally.}

\begin{abstract}
In this study, we use recent stance detection methods to study the stance (for, against or neutral) of statements in official information booklets for voters. Our main goal is to answer the fundamental question: are topics to be voted on presented in a neutral way?
To this end, we first train and compare several models for stance detection on a large dataset about Swiss politics. We find that fine-tuning an M-BERT model leads to the best accuracy. We then use our best model to analyze the stance of utterances extracted from the Swiss federal voting booklet concerning the Swiss popular votes of September 2022, which is the main goal of this project.
We evaluated the models in both a multilingual as well as a monolingual context for German, French, and Italian. Our analysis shows that some issues are heavily favored while others are more balanced, and that the results are largely consistent across languages. 
Our findings have implications for the editorial process of future voting booklets and the design of better automated systems for analyzing political discourse. The data and code accompanying this paper are available at
\href{https://github.com/ZurichNLP/voting-booklet-bias}{https://github.com/ZurichNLP/voting-booklet-bias}.

\end{abstract}


\maketitle

\section{Introduction}
\label{sec:introduction}


In this work, we investigate the neutrality of statements presented to voters in official information booklets by using state-of-the-art stance detection methods.

Many democratic countries in the world conduct popular votes on important decisions where the majority of the population is involved. As an example, Switzerland routinely holds popular votes several times each year, and each popular vote is comprised of several independent topics, such as proposed changes to current law\footnote{\url{https://www.admin.ch/gov/en/start/documentation/votes.html}}. Constituents are given a choice to either vote for or against each topic. The Swiss government issues a \textit{voting booklet}, an explanatory brochure, to inform citizens about upcoming popular votes.

We argue that it is in the best interest of citizens that governments issue voting information in a neutral manner. Especially in countries like Switzerland that have a strong democratic governance model, official communication should be presented without favouring one particular  point of view. We assume that Swiss voters ultimately form their opinions based on the information given in the booklet. Our paper builds on this premise and poses the question: are statements in voting information indeed neutral?

To examine this question, we propose to use recent stance detection methods that are widely used in Natural Language Processing (NLP). Stance detection is the task of automatically detecting an author's stance towards a subject (called the \textit{target}) \citep{10.1145/3369026,hardalov-etal-2022-survey,alturayeif2023systematic}. A \textit{stance} is understood as a writer's attitude towards, evaluation of or alignment with a particular subject \citep{du2007stance,kockelman2004stance}. See Table~\ref{tab:stances} for an illustration of stance prediction with examples.

We study the stances of one particular dataset, namely voting booklets from Switzerland's popular votes in September 2022. In September 2022, the popular vote comprised four separate topics (see Section~\ref{subsec:the_swiss_federal_voting_booklet} for more details). The dataset is multilingual since the voting booklet is available in all four official languages of Switzerland: German, French, Italian and Romansh. This allows us to compare the stances towards four different topics, across three\footnote{We exclude Romansh since it is not among the languages M-BERT is pretrained on. See \url{https://github.com/google-research/bert/blob/master/multilingual.md}} languages.

In order to create a stance prediction model for this specific dataset, we build on the work of \citet{vamvas2020xstance} who have shown that a pretrained multilingual BERT (M-BERT) model \citep{devlin2018bert} can be fine-tuned for stance prediction on Swiss politics data. We reproduce their experiments, including comparisons of M-BERT to other stance prediction methods, and confirm that an M-BERT model indeed is the highest-performing approach.

Our main finding is that the statements for two out of four popular votes heavily favour one voting outcome. Moreover, our multilingual analysis shows that this finding is consistent across three different languages.
These findings invite more analysis of the kind of language commonly found in voter information, and a potential implication is that future voting booklets should feature more neutral statements.

\begin{table*}
    \caption{Illustration of stance detection, using an example from the Swiss voting booklet of the September 2022 popular vote (introduced in Section~\ref{subsec:the_swiss_federal_voting_booklet}). A stance detection model predicts the stance of a \textit{statement} towards a \textit{target}. The target can be implicit, embedded into a natural language sentence. In the example below, the real target (the \textit{Factory Farming initiative}) is mentioned implicitly in a question. We also include the stances predicted by our model (explained in Section~\ref{sec:model_training}).
    English translations were generated with DeepL. DE=German, FR=French, IT=Italian.}
    \label{tab:stances}
  \begin{tabular}{ p{11em} p{11em} p{11em} p{11em} }
    \toprule
     \textbf{Question (target)} & \textbf{DE} & \textbf{FR} & \textbf{IT} \\
    \midrule
    \textit{Do you want to accept the popular initiative "No factory farming in Switzerland (Factory Farming initiative)"?} & Wollen Sie die Volksinitiative «Keine Massentierhaltung in der Schweiz (Massentierhaltungsinitiative)» annehmen? &
    Acceptez-vous l’initiative populaire «Non à l’élevage intensif en Suisse (initiative sur l’élevage intensif)» ? &
    Volete accettare l’iniziativa popolare «No all’allevamento intensivo in Svizzera (Iniziativa sull’allevamento intensivo)»? \\\\
     \textbf{Statement} & & & \\
     \midrule
     \textit{Switzerland has one of the strictest laws in the world for the protection of animals. The dignity and welfare of animals are protected, regardless of how many animals are kept in one place.} & Die Schweiz hat eines der weltweit strengsten Gesetze zum Schutz der Tiere. Würde und Wohlergehen von Tieren sind geschützt, unabhängig  davon, wie viele Tiere an einem Ort gehalten werden. &
    La loi suisse sur la protection des animaux est l’une des plus strictes au monde. La dignité et le bien-être des animaux sont protégés, indépendamment du nombre d’animaux détenus au même endroit. &
    La Svizzera dispone di una legge sulla protezione degli animali fra le più severe al mondo. La dignità e il benessere degli animali sono tutelati, indipendentemente dal numero di capi detenuti in un allevamento. \\
    \textbf{Detected stance} & & & \\
    \midrule
      & \textit{NEUTRAL} & \textit{NEUTRAL} & \textit{NEUTRAL} \\
     \bottomrule
  \end{tabular}
\end{table*}

\section{Related work}
\label{sec:related_work}

\paragraph{Stance detection}

Stance detection is the task of automatically detecting an author's stance towards a subject (called the \textit{target}) \citep{10.1145/3369026,hardalov-etal-2022-survey,alturayeif2023systematic}. It is often framed as a binary classification problem with the classes ``for'' and ``against'' or including ``neutral'' as an additional label.

\paragraph{Political analysis and NLP}

NLP methods have become a well-established tool in the field of political science for studying phenomena that are of interest to the discipline \citep{wilkerson2017large,chatsiou2020deep, glavas-etal-2019-computational,politicalnlp-2022-lrec}. Examples for widely used methods are sentiment analysis, misinformation detection and stance detection, which are used to analyze various forms of textual data, including political speeches \citep{guerini2013new,barbaresi-2018-corpus}, social media posts \citep{mohammad-etal-2016-semeval,kawintiranon-singh-2022-polibertweet} and news articles \citep{chen-etal-2020-analyzing}.

\paragraph{Stance prediction on political data}

Recent publications on stance prediction for political data (such as speeches, debates, political party manifestos or tweets) include \citet{lai-muli-stances}, \citet{vamvas2020xstance}, \citet{barriere-etal-2022-debating} and \citet{bergam-etal-2022-legal}. Among those, \citet{vamvas2020xstance} is the most similar to our work. \citeauthor{vamvas2020xstance} introduced a dataset called \textit{x-stance} for multi-target stance detection, consisting of 67k comments of electoral candidates in Switzerland taken from a voting information platform. The corpus is multilingual, with comments available in German, French and Italian. They then trained a multi-target stance detection model based on M-BERT \citep{devlin2018bert} that predicts binary labels (``for'' or ``against'').

In order to analyze the stances in our own dataset (a voting booklet introduced in Section~\ref{subsec:the_swiss_federal_voting_booklet}), we reproduce the models of \citeauthor{vamvas2020xstance}, training these models on the dataset they published.

\section{Datasets}
\label{sec:data_sets}

This section describes in more detail the datasets we use in our work. We use the \textit{x-stance} dataset by \citet{vamvas2020xstance} (introduced in Section~\ref{subsec:swiss_voting_data}) to train stance prediction models. We use a Swiss voting booklet (introduced in Section~\ref{subsec:the_swiss_federal_voting_booklet}) for analysis only (not for training any model).

\subsection{The \textit{x-stance} dataset}
\label{subsec:swiss_voting_data}

\citet{vamvas2020xstance} collected data from Smartvote\footnote{\url{https://smartvote.ch}}, a free online platform dedicated to voting and political views in Switzerland. Smartvote allows voters to determine the political views of incumbent politicians or electoral candidates on a variety of topics. The resulting dataset, called \textit{x-stance}, covers more than 150 political questions and 67k comments from political figures. The corpus is multilingual, with comments available in German, French and Italian -- and freely available.

We use \textit{x-stance} to train multilingual, multi-target stance detection models, following closely the methodology of \citet{vamvas2020xstance}. The reasons for training our own models in this project are, on the one hand, that we do not have access to the original model and, on the other hand, that we want to ensure the reproducibility of the chosen models. The model training is explained further in Section~\ref{sec:model_training}.

\subsection{The Swiss federal voting booklet}
\label{subsec:the_swiss_federal_voting_booklet}

\paragraph{General nature of Swiss voting booklets}

The Swiss federal voting booklets are official documents issued by the government, with the main goal of informing voters about upcoming popular votes. The booklets are published in each national language -- German, French, Italian and Romansh. The booklets convey the points of view of the parliament, federal council and initiative committee (political entity that is advocating for a change to current law, initiating the process of a popular vote). Even though the views of several political actors are presented, the booklets aim for a balanced presentation overall, so as not to polarize potentially uninformed voters. According to its publisher, the Federal Chancellery, the booklet is designed to present opinions from both sides evenly \citep{bk-comm}.

These characteristics make for a great data source for stance detection. They allow to validate whether the description of each initiative (issue to vote on) provides an overall balanced view, that is, a roughly equal amount of negative and positive stances. These findings can then be compared across languages and issues. 

Conveniently, each of the issues in a booklet are contained in a separate chapter, described as a series of paragraphs of approximately equal length. The core content usually consists of a short summary, an introduction, and a few pages containing arguments in favor and against the issue.

\paragraph{Particular dataset used in our study}

To analyze stances in official voting information we extracted data from the voting booklet issued for Switzerland's popular vote in September 2022. The popular vote consisted of four separate issues to be voted on (we include abbreviations we use later in the text):

\begin{itemize}
    \item The Factory Farming Initiative (FFI)
    \item First reform of the old-age and survivors's insurance, on VAT (OASI-1)
    \item Second reform of the old-age and survivors's insurance, on retirement age (OASI-2)
    \item Amendment to the Federal Act on Withholding Tax (FAWT)
\end{itemize}

We downloaded the latest version of the booklets (June 15, 2022)\footnote{See \url{https://www.admin.ch/gov/de/start/dokumentation/abstimmungen/20220925.html} for the German version} as a PDF and extracted text data from them by hand. We kept all the text in German, French and Italian, and for each statement recorded which issue (FFI, OASI-1, OASI-2 or FAWT) it belongs to. See Table~\ref{tab:stances} for an example of the resulting dataset.
Table~\ref{tab:distr} summarizes the distribution of the statements we extracted from the latest German edition of the booklet. The distribution over issues is quite even, slightly more space is dedicated to the explanations of the OASI reforms 1 and 2, both of which were described with approximately $25\%$ more paragraphs. The other languages (French and Italian), although not shown, show a similar distribution.

\begin{table*}
    \caption{The distribution of statements in the German voting booklet containing a total of n = 407 statements across all issues. Each row represents one of the four political reforms along with the absolute (relative) amount of statements describing it and the average number of characters per statement. While the booklet consists of  comparably more statements describing the two OASI reforms, they were roughly of the same length across all reforms. The French and Italian voting booklets display similar distributions.}
    \label{tab:distr}
  \begin{tabular}{ lrr }
    \toprule
     Reform & \textbf{Number of statements (\%)} & \textbf{Average length of one statement (characters)} \\
    \midrule
     Factory Farming & 88 (21.6\%) & 432 \\
     OASI Reform 1 & 116 (28.5\%) & 459 \\
     OASI Reform 2 & 115 (28.3\%) & 470 \\
     Withholding Tax & 88 (21.6\%) & 396 \\
     \midrule
     Total & 407 (100\%) &  \\
     \bottomrule
  \end{tabular}
\end{table*}

\section{Model training}
\label{sec:model_training}

We train several models for multilingual, multi-target stance detection, using the \emph{x-stance corpus} (see Section~\ref{subsec:swiss_voting_data}) as training data and attempting to reproduce the results reported in \citet{vamvas2020xstance}. We decided to follow \citet{vamvas2020xstance} because they propose a multilingual and multi-target approach, which is also required for our analysis (three languages and 4 separate targets). Reproducing their experiments was necessary because their models are not available.

\subsection{Reproduction of models and results from earlier work}
\label{subsec:reproduction}

Following \citet{vamvas2020xstance}, we trained both a fastText \citep{joulin2016bag} and a fine-tuned M-BERT \citep{devlin2018bert} model on the \textit{x-stance} dataset. fastText is a lightweight linear classifier based on character and word embeddings \cite{joulin2016bag}. M-BERT is a Transformer-based model pre-trained on large unlabeled corpora in 104 languages which was shown to enable cross-lingual transfer for NLP tasks \citep{devlin2018bert}. The flexibility of applying transfer learning to a pre-trained BERT model allows us to perform our experiments without a large computational overhead. Specifically, \citet{devlin2018bert} state that M-BERT can be fine-tuned for a wide range of tasks with just one additional output layer and without substantial task-specific architecture modifications.

If not noted otherwise, all preprocessing, training and evaluation decisions are reproduced exactly from \citet{vamvas2020xstance}.
\textit{x-stance} is split into separate train, validation and test datasets. For German and French, the train/dev/test sets contain roughly 85\%/7.5\%/7.5\% of all statements in that language, respectively. Italian samples only occur in the test set, to test zero-shot cross-lingual transfer. For training we use the same hyperparameters as \citet{vamvas2020xstance}, given that they performed well on the Smartvote data which is similar to our voting booklet data (see Section~\ref{sec:data_sets} for an explanation of the difference).

Please note that we are not aiming to improve over the results of \citet{vamvas2020xstance}. Our goal is to reproduce their work as closely as possible, in order to use the resulting model to analyze voter information data.

\subsection{Additional baseline models}
\label{subsec:additional_baseline_models}

To provide additional baselines for benchmark comparisons, we additionally trained two popular linear classifiers on the training data, namely a ridge regression and support vector machine model. We use Scikit-learn \citep{scikit-learn} to implement and train both models.

As the only preprocessing step, we transform the text data to continuous features using a feature hashing method\footnote{\url{https://scikit-learn.org/stable/modules/generated/sklearn.feature_extraction.FeatureHasher.html}}. We use the default hyperparameters for training both classifiers.

\subsection{Stance prediction performance of trained models}
\label{subsec:stance_prediction_performance}

The results of the model training and evaluation are shown in Table~\ref{tab:baseline}, where the macro-average of the F1-scores for in favor and against are shown. We also split our evaluation into separate scores for \textit{intra-target}, \textit{cross-question} and \textit{cross-topic} partitions of \textit{x-stance}, following \citet{vamvas2020xstance}. \textit{intra-target} tests the model performance on known questions and topics that were seen during training. \textit{cross-question} and \textit{cross-topic} are designed to test the performance on unseen kinds of data, where either the type of question or topic are held-out during training.

We note that the reproduction of the model training by \citet{vamvas2020xstance} was successful, resulting in similar scores for all the different metrics. All our reported F1 scores in Table~\ref{tab:baseline} are within a three-percentage-point range of the original values. This validates the original study by \citet{vamvas2020xstance} and we confirm that their experiments are sound.

While all models performed quite well in our benchmark for the \textit{intra-target} partition, there is a significant gap in the scores for both the cross-question and cross-topic partitions.  Inspecting the scores more closely reveals that the generalization ability of the fastText and the M-BERT models is higher than our linear classifier baselines. Comparing only fastText and M-BERT, the evaluation clearly demonstrates that M-BERT is superior, scoring the highest in every aspect of the benchmark. Based on these results we concluded that applying the trained M-BERT model to our voting booklet data is most suitable.

\begin{table*}[t]
\caption{F1-scores of baselines, fastText and M-BERT models for German (DE) and French (FR) test set samples, including their harmonic mean as a third metric. \textit{intra-target} = performance on known questions and topics that were seen during training. \textit{cross-question} and \textit{cross-topic} = performance on unseen kinds of data, where either the type of question or topic are held-out during training. The second group of results are scores reported by \citet{vamvas2020xstance} that we are aiming to reproduce.}
\centering
\begin{tabular}{ p{2cm} p{0.8cm} p{0.8cm} p{1cm} p{0.8cm} p{0.8cm} p{1cm} p{0.8cm} p{0.8cm} p{1cm} }
 \hline
 & \multicolumn{3}{c}{\textbf{Intra-target}} & \multicolumn{3}{c}{\textbf{Cross-question}} & \multicolumn{3}{c}{\textbf{Cross-topic}} \\
 & DE & FR & Mean & DE & FR & Mean & DE & FR & Mean\\
 \hline
Ridge (baseline) & 61.22 & 67.01 & 64.12 & 37.49 & 40.36 & 38.93 & 34.70 & 47.66 & 41.18 \\
SVM (baseline) & 61.49 & 67.05 & 64.27 & 37.49 & 40.13 & 38.81 & 34.70 & 45.34 & 40.02 \\
\midrule
fastText \citep{vamvas2020xstance} & 69.90 & 71.20 & 70.50 & 62.00 & 65.60 & 63.70 & 63.10 & 65.50 & 64.30 \\
M-BERT \citep{vamvas2020xstance} & 76.80 & 76.60 & 76.60 & 68.50 & 68.40 & 68.40 & 68.90 & 70.90 & 69.90 \\
\midrule
fastText (ours) & 69.37 & 71.45 & 70.41 & 62.07 & 62.70 & 62.39 & 62.83 & 63.37 & 63.10 \\
M-BERT (ours) & 76.57 & 78.13 & 77.35 & 66.72 & 68.88 & 67.80 & 68.00 & 69.37 & 68.69 \\
 \hline
\end{tabular}
\label{tab:baseline}
\end{table*}

\section{Heuristic for neutral stances}
\label{sec:neutral_stances}

\textit{x-stance} is a dataset for binary stance prediction, lacking a third class that is neither \textit{for} or \textit{against}. In this section we argue for this third type of label and propose a way to construct it without re-training our models.

Applying the stance detection model to each sample yielded both a label (\textit{against} or \textit{in favor}) and a two-dimensional vector with the normalized probabilities $p \in (0, 1)$ of each label. From this vector, the "favor" probability has been extracted, which serves as the main measurement for our further analyses.

However, in the context of political communication, it may be a good idea to allow a text to have \textit{neutral} stance. \citet{mohammad-etal-2016-semeval} argue that the lack of evidence for "in favor" or "against" does not necessarily imply a neutral stance, it may simply mean no stance can be detected. \textit{Neither} might be a more accurate label but for simplicity, we will use the term "neutral" in this paper \citep{lai-muli-stances}.

Given that our stance detection model is unable to explicitly classify neutral stances, we propose a heuristic to detect the absence of a clear stance. 
We take a probabilistic view, as follows: we label an absence explicitly as a neutral stance based on the standard deviation of favor probabilities for a specific issue $t$. A corresponding statement $s$ with favor probability $p_f(s)$ is assigned the neutral label if $p_f(s) - \frac{1}{2} \in [-\sigma_t, \sigma_t]$. The constant normalization factor $\frac{1}{2}$ comes from the sigmoid output probabilities of the original model: If both labels (favor and against) were equally likely, then $p_f(s)=p_a(s)=\frac{1}{2}$. Statements are assigned a neutral label if $p_f(s)$ is within one standard deviation. This statistically motivated approach introduces context-dependence: Whether or not a statement exhibits a neutral stance depends on the strength of stances of all other statements addressing the same target.  

\section{Results and discussion}
\label{sec:results_and_discussion}

We apply the M-BERT model trained in Section~\ref{sec:model_training} to the voting booklet data introduced in Section~\ref{subsec:the_swiss_federal_voting_booklet}.

\paragraph{A priori expectation}

As a general rule, each issue in a voting booklet should be presented in a neutral and balanced manner (see Sections~\ref{sec:introduction} and~\ref{subsec:the_swiss_federal_voting_booklet}). We expected most statements to be neutral, and the remaining statements (\textit{for} and \textit{against}) to be roughly evenly distributed, in other words, an equal amount of statements for each stance.

\begin{figure*}
\centering
\includegraphics[width=\linewidth]{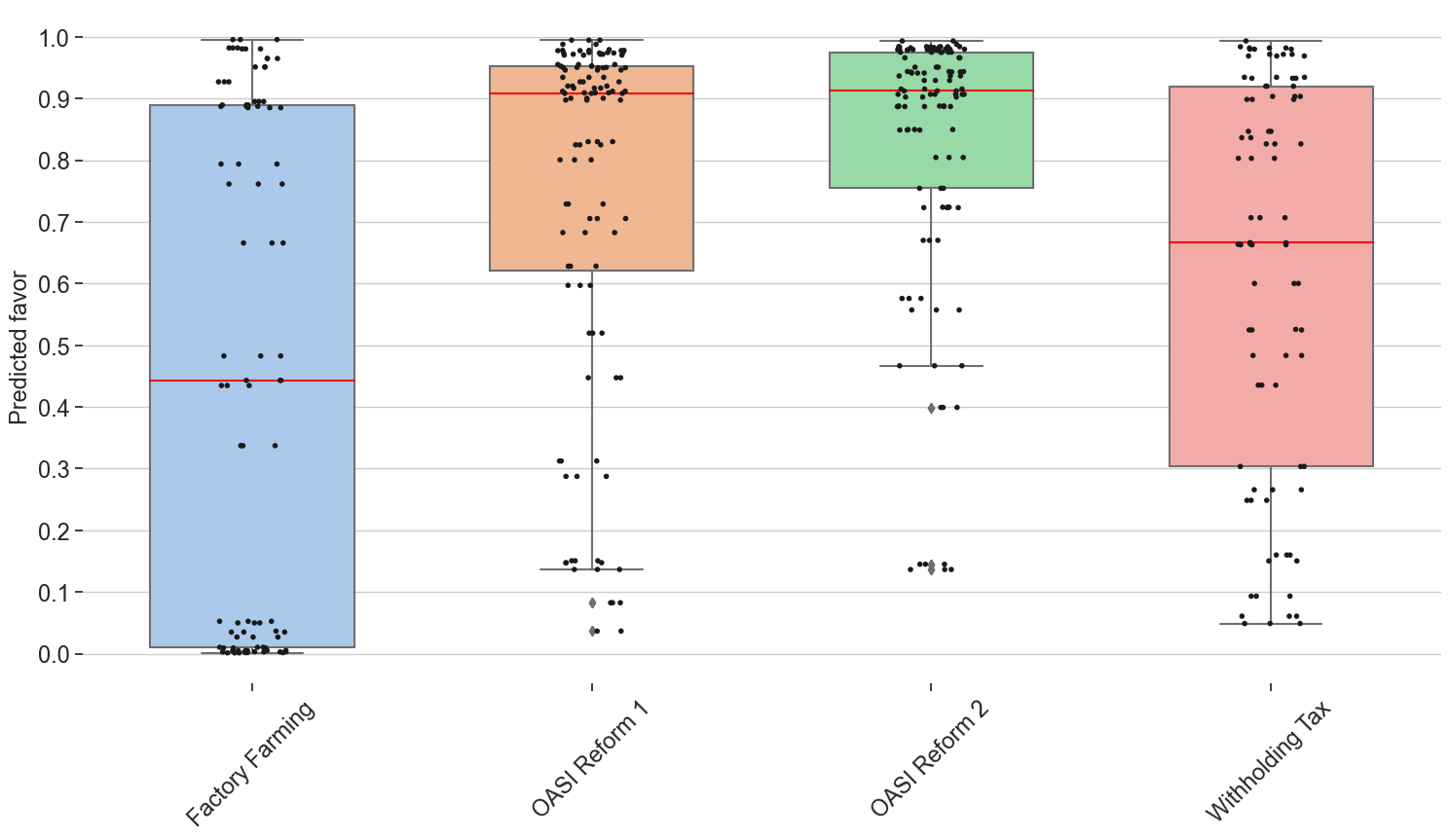}
\caption{Distribution of stance predictions for each issue of the popular vote. Results are aggregated across all languages (German, French and Italian). The y-axis denotes the probability of the ``for'' label, indicating that a statement supports the target issue. The whiskers extend to the interquartile range. The red line marks the median.}
\label{fig:stance-distribution}
\end{figure*}

\paragraph{Predicted favour} As a first way to visualize the prediction results, Figure~\ref{fig:stance-distribution} shows the distribution of stance predictions for each issue of the popular vote. The graph shows the probability of the ``for'' label, which we refer to as ``predicted favour''. Results are aggregated across all languages (German, French and Italian). Additional plots (with similar trends) for the monolingual evaluations can be found in Appendix~\ref{app:boxplots-languages}.

Figure~\ref{fig:stance-distribution} demonstrates that the Factory Farming initiative is the only part of the voting booklet where statements are balanced: the median predicted favour is close to 0.5. All other issues are described by statements that heavily favour them, potentially offering a one-sided and biased point of view. This trend is most extreme for the two OASI reforms where the median predicted favour is above 0.9.

\paragraph{Including neutral stances}

\begin{figure*}
\centering
\includegraphics[width=\linewidth]{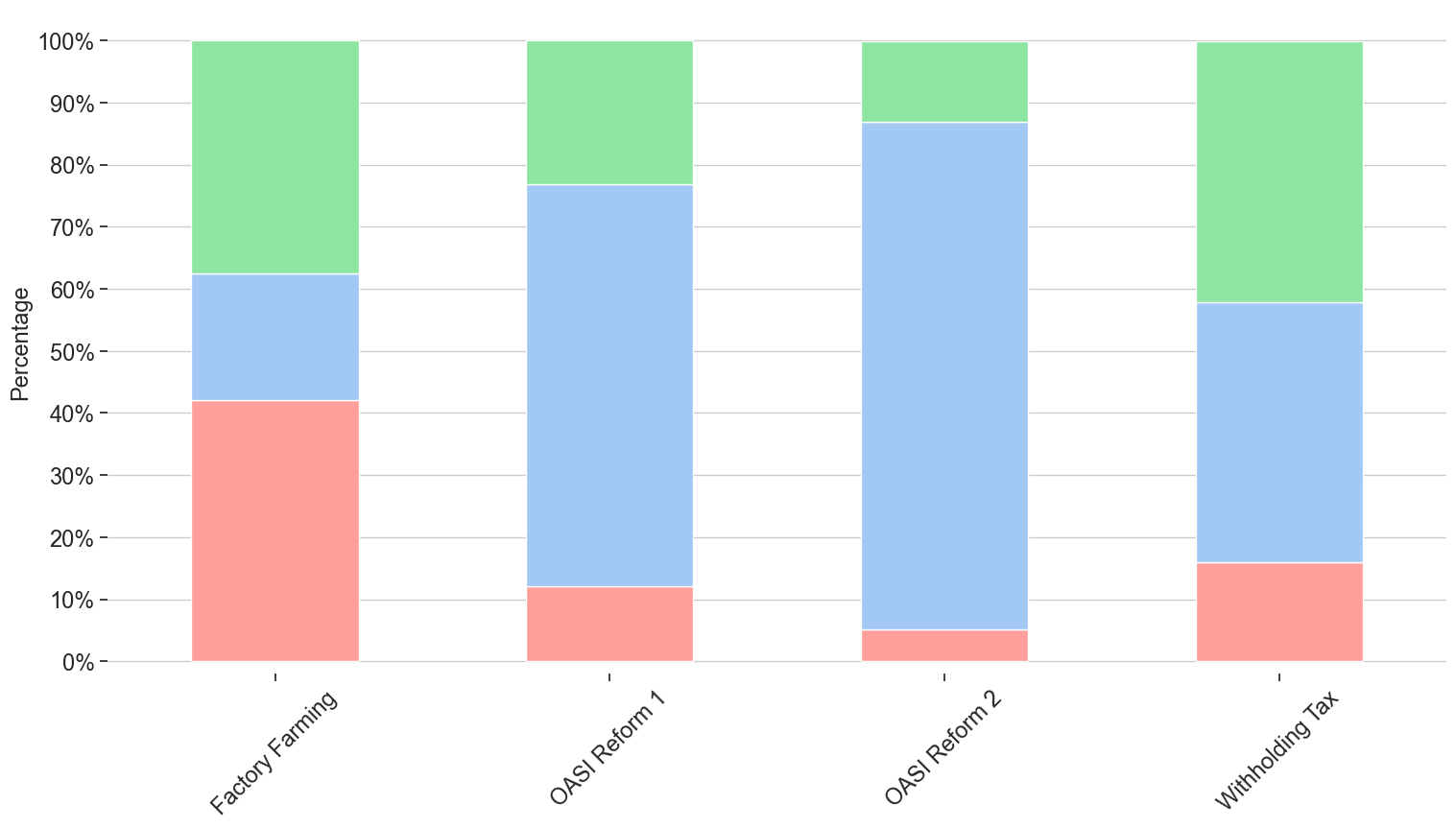}
\caption{Distribution of stance labels for each issue of the popular vote. The graph shows a three-way classification between ``for'', ``against'' and ``neutral'' (a heuristic, synthetic label). Results are aggregated across all languages (German, French and Italian). 
Red is against, blue is neutral and green is favor.}
\label{fig:label-distribution}
\end{figure*}

As a second way of reporting results, Figure~\ref{fig:label-distribution} incorporates heuristic labels for ``neutral stances'' (explained in Section~\ref{sec:neutral_stances}). The figure shows the relative percentages of stance labels for each issue of the popular vote. The graph shows a three-way classification between ``for'', ``against'' and ``neutral'' (the synthetic label not in fact produced directly by the model). Results are aggregated across all languages (German, French and Italian).

Adding in our heuristic for which predictions could in reality be neutral stances shows that most of the statements concerning the two OASI reforms appear neutral, while both still show a higher percentage of favoring statements than disfavoring ones. The same is true for the Witholding Tax, and only the information about the Factory Farming initiative is composed of a roughly equal amount of statements in favor and against the initiative. 

To summarize, only the statements on the Factory Farming initiative show equal distribution of stances.
For the three other issues being voted on, all statements taken together tend to favour a legal reform (as opposed to favouring the status quo).

\subsection{Further discussion}

\paragraph{Similar patterns for individual languages}

Our analysis shows similar trends for individual languages (see Appendix~\ref{app:boxplots-languages}). In the French and Italian versions of the booklet, the two OASI reforms were slightly more balanced while the Factory Farming initiative appears almost identical to the German equivalent. A possible reason for this is the fact that our M-BERT model was trained on a much larger sample of German data from \textit{x-stance} \citep{vamvas2020xstance}. Inevitably, predictions for other languages may result in more uncertainty, which manifests itself in values being dragged towards the center of the normalized probability range.

\paragraph{Special legal status of OASI reforms}

It is important to note that the two OASI reforms were tied together. That is, both partial reforms needed to be accepted by the Swiss population in order for the overarching OASI reform to take effect. This may explain the similar statement distribution in Table~\ref{tab:distr} that indicates a focus on the two reforms. Likewise, the stance distributions in Figure~\ref{fig:label-distribution} show a similar tendency for both OASI reforms. 

\paragraph{Voting outcome}

On September 25, 2022, the Swiss population voted in favor of the two OASI reforms and against both the Factory Farming and Withholding Tax initiative. Directly relating our findings to the results of the votes would be a huge leap of faith. Instead, we assume that the linguistic characteristics of the data accounts for most of the variance. In the context of the German booklet, the ten most disfavoring statements over all issues can be found in the Factory Farming initiative. And while the most favorable paragraphs are distributed somewhat evenly across all issues, they are generally shorter in length than their counterparts and may therefore be easier to read.

\paragraph{Neutral stances and emotionally charged language}

The political issues we examined are very different in their nature, ranging from taxes and insurance to farming. They are described with a very different vocabulary consisting of either more abstract financial or legal terms (e.g. Withholding Tax) or resembling everyday terminology more (e.g. Factory Farming) terms. Intuitively, it seems easier to express a stance towards emotionally charged words such as "pesticides" or "animal welfare". In contrast, the descriptions of the other reforms contain more factual terms such as "tax rate" or "pension fund" and hence do not exhibit equally strong stances. The derivation of a context-dependent neutral label helped to put these linguistic characteristics into perspective, which is corroborated by large proportion of neutral labels found in the two OASI reforms in Table~\ref{tab:distr}.

\section{Conclusion}
\label{sec:conclusion}

In this work we investigated whether statements in a Swiss voting booklet are presented in a neutral way, using stance detection methods. To this end we trained stance detection models on a dataset of Swiss political language, reproducing earlier work.

All four issues we examined were given roughly the same space and explained with the same number of statements (see Table~\ref{tab:distr}). Yet, we found that the voting booklets, across all languages, clearly favour two our of four issues. Statements tended to be in favor of the two OASI reforms and features somewhat balanced arguments both in favor and against the other two initiatives (see Figure~\ref{fig:stance-distribution}).

\section{Future work}
\label{sec:future_work}

We believe that future work could analyze linguistic characteristics, examine a larger dataset of voting booklets or explore the relationship between most frequent stances and voting outcomes.

\paragraph{Analysis of linguistic characteristics}

We reported on resulting stance distributions without investigating the linguistic characteristics and differences between issues. Future work could investigate this in more detail and study the specific linguistic characteristics in texts that result in particular stances. In particular, an analysis of topical and rhetorical framing could be insightful \cite{yu-2023-towards}.  

\paragraph{Larger dataset}

Creating and analyzing a larger dataset of more voting booklets could strengthen our empirical results. In addition, it would enable us to add a diachronic dimension and show how stance trends develop over time. Much more historical data for Swiss popular votes would be available to digitize and analyze.

\paragraph{Relationship to voting outcomes}

In a similar vein, the outcomes of all past popular votes could be analyzed together with predicted stances, to test for potential correlations between them.

\begin{acknowledgments}
Mathias Müller was funded by the EU Horizon 2020 project EASIER (grant agreement no. 101016982).\\
We thank Marc Egli, MSc Statistics ETH, for the insightful discussions concerning neutral stance heuristics and Nicolas Koch, MSc Statistics ETH, for clarifications regarding the statistical significance of differences in the reported accuracy scores.
\end{acknowledgments}

\bibliography{sample-ceur}

\begin{thebibliography}{23}
\expandafter\ifx\csname natexlab\endcsname\relax\def\natexlab#1{#1}\fi
\providecommand{\url}[1]{\texttt{#1}}
\providecommand{\href}[2]{#2}
\providecommand{\path}[1]{#1}
\providecommand{\DOIprefix}{doi:}
\providecommand{\ArXivprefix}{arXiv:}
\providecommand{\URLprefix}{URL: }
\providecommand{\Pubmedprefix}{pmid:}
\providecommand{\doi}[1]{\href{http://dx.doi.org/#1}{\path{#1}}}
\providecommand{\Pubmed}[1]{\href{pmid:#1}{\path{#1}}}
\providecommand{\bibinfo}[2]{#2}
\ifx\xfnm\relax \def\xfnm[#1]{\unskip,\space#1}\fi
\bibitem[{K\"{u}\c{c}\"{u}k and Can(2020)}]{10.1145/3369026}
\bibinfo{author}{D.~K\"{u}\c{c}\"{u}k}, \bibinfo{author}{F.~Can},
\newblock \bibinfo{title}{Stance detection: A survey},
\newblock \bibinfo{journal}{ACM Comput. Surv.} \bibinfo{volume}{53}
  (\bibinfo{year}{2020}). \URLprefix \url{https://doi.org/10.1145/3369026}.
  \DOIprefix\doi{10.1145/3369026}.
\bibitem[{Hardalov et~al.(2022)Hardalov, Arora, Nakov, and
  Augenstein}]{hardalov-etal-2022-survey}
\bibinfo{author}{M.~Hardalov}, \bibinfo{author}{A.~Arora},
  \bibinfo{author}{P.~Nakov}, \bibinfo{author}{I.~Augenstein},
\newblock \bibinfo{title}{A survey on stance detection for mis- and
  disinformation identification},
\newblock in: \bibinfo{booktitle}{Findings of the Association for Computational
  Linguistics: NAACL 2022}, \bibinfo{publisher}{Association for Computational
  Linguistics}, \bibinfo{address}{Seattle, United States},
  \bibinfo{year}{2022}, pp. \bibinfo{pages}{1259--1277}. \URLprefix
  \url{https://aclanthology.org/2022.findings-naacl.94}.
  \DOIprefix\doi{10.18653/v1/2022.findings-naacl.94}.
\bibitem[{Alturayeif et~al.(2023)Alturayeif, Luqman, and
  Ahmed}]{alturayeif2023systematic}
\bibinfo{author}{N.~Alturayeif}, \bibinfo{author}{H.~Luqman},
  \bibinfo{author}{M.~Ahmed},
\newblock \bibinfo{title}{A systematic review of machine learning techniques
  for stance detection and its applications},
\newblock \bibinfo{journal}{Neural Computing and Applications}
  (\bibinfo{year}{2023}) \bibinfo{pages}{1--32}.
\bibitem[{Du~Bois(2007)}]{du2007stance}
\bibinfo{author}{J.~W. Du~Bois},
\newblock \bibinfo{title}{The stance triangle},
\newblock \bibinfo{journal}{Stancetaking in discourse: Subjectivity,
  evaluation, interaction} \bibinfo{volume}{164} (\bibinfo{year}{2007})
  \bibinfo{pages}{139--182}.
\bibitem[{Kockelman(2004)}]{kockelman2004stance}
\bibinfo{author}{P.~Kockelman},
\newblock \bibinfo{title}{Stance and subjectivity},
\newblock \bibinfo{journal}{Journal of Linguistic Anthropology}
  \bibinfo{volume}{14} (\bibinfo{year}{2004}) \bibinfo{pages}{127--150}.
\bibitem[{Vamvas and Sennrich(2020)}]{vamvas2020xstance}
\bibinfo{author}{J.~Vamvas}, \bibinfo{author}{R.~Sennrich},
\newblock \bibinfo{title}{{X-Stance}: A multilingual multi-target dataset for
  stance detection},
\newblock in: \bibinfo{booktitle}{Proceedings of the 5th Swiss Text Analytics
  Conference (SwissText) \& 16th Conference on Natural Language Processing
  (KONVENS)}, \bibinfo{address}{Zurich, Switzerland}, \bibinfo{year}{2020}.
  \URLprefix \url{http://ceur-ws.org/Vol-2624/paper9.pdf}.
\bibitem[{Devlin et~al.(2019)Devlin, Chang, Lee, and
  Toutanova}]{devlin2018bert}
\bibinfo{author}{J.~Devlin}, \bibinfo{author}{M.-W. Chang},
  \bibinfo{author}{K.~Lee}, \bibinfo{author}{K.~Toutanova},
\newblock \bibinfo{title}{{BERT}: Pre-training of deep bidirectional
  transformers for language understanding},
\newblock in: \bibinfo{booktitle}{Proceedings of the 2019 Conference of the
  North {A}merican Chapter of the Association for Computational Linguistics:
  Human Language Technologies, Volume 1 (Long and Short Papers)},
  \bibinfo{publisher}{Association for Computational Linguistics},
  \bibinfo{address}{Minneapolis, Minnesota}, \bibinfo{year}{2019}, pp.
  \bibinfo{pages}{4171--4186}. \URLprefix
  \url{https://aclanthology.org/N19-1423}.
  \DOIprefix\doi{10.18653/v1/N19-1423}.
\bibitem[{Wilkerson and Casas(2017)}]{wilkerson2017large}
\bibinfo{author}{J.~Wilkerson}, \bibinfo{author}{A.~Casas},
\newblock \bibinfo{title}{Large-scale computerized text analysis in political
  science: Opportunities and challenges},
\newblock \bibinfo{journal}{Annual Review of Political Science}
  \bibinfo{volume}{20} (\bibinfo{year}{2017}) \bibinfo{pages}{529--544}.
\bibitem[{Chatsiou and Mikhaylov(2020)}]{chatsiou2020deep}
\bibinfo{author}{K.~Chatsiou}, \bibinfo{author}{S.~J. Mikhaylov},
\newblock \bibinfo{title}{Deep learning for political science},
\newblock \bibinfo{journal}{arXiv preprint arXiv:2005.06540}
  (\bibinfo{year}{2020}).
\bibitem[{Glava{\v{s}} et~al.(2019)Glava{\v{s}}, Nanni, and
  Ponzetto}]{glavas-etal-2019-computational}
\bibinfo{author}{G.~Glava{\v{s}}}, \bibinfo{author}{F.~Nanni},
  \bibinfo{author}{S.~P. Ponzetto},
\newblock \bibinfo{title}{Computational analysis of political texts: Bridging
  research efforts across communities},
\newblock in: \bibinfo{booktitle}{Proceedings of the 57th Annual Meeting of the
  Association for Computational Linguistics: Tutorial Abstracts},
  \bibinfo{publisher}{Association for Computational Linguistics},
  \bibinfo{address}{Florence, Italy}, \bibinfo{year}{2019}, pp.
  \bibinfo{pages}{18--23}. \URLprefix
  \url{https://www.aclweb.org/anthology/P19-4004}.
  \DOIprefix\doi{10.18653/v1/P19-4004}.
\bibitem[{Afli et~al.(2022)Afli, Alam, Bouamor, Casagran, Boland, and
  Ghannay}]{politicalnlp-2022-lrec}
\bibinfo{editor}{H.~Afli}, \bibinfo{editor}{M.~Alam},
  \bibinfo{editor}{H.~Bouamor}, \bibinfo{editor}{C.~B. Casagran},
  \bibinfo{editor}{C.~Boland}, \bibinfo{editor}{S.~Ghannay} (Eds.),
  \bibinfo{title}{Proceedings of the LREC 2022 workshop on Natural Language
  Processing for Political Sciences}, \bibinfo{publisher}{European Language
  Resources Association}, \bibinfo{address}{Marseille, France},
  \bibinfo{year}{2022}. \URLprefix
  \url{https://aclanthology.org/2022.politicalnlp-1.0}.
\bibitem[{Guerini et~al.(2013)Guerini, Giampiccolo, Moretti, Sprugnoli, and
  Strapparava}]{guerini2013new}
\bibinfo{author}{M.~Guerini}, \bibinfo{author}{D.~Giampiccolo},
  \bibinfo{author}{G.~Moretti}, \bibinfo{author}{R.~Sprugnoli},
  \bibinfo{author}{C.~Strapparava},
\newblock \bibinfo{title}{The new release of corps: A corpus of political
  speeches annotated with audience reactions},
\newblock in: \bibinfo{booktitle}{Multimodal Communication in Political Speech.
  Shaping Minds and Social Action: International Workshop, Political Speech
  2010, Rome, Italy, November 10-12, 2010, Revised Selected Papers},
  \bibinfo{organization}{Springer}, \bibinfo{year}{2013}, pp.
  \bibinfo{pages}{86--98}.
\bibitem[{Barbaresi(2018)}]{barbaresi-2018-corpus}
\bibinfo{author}{A.~Barbaresi},
\newblock \bibinfo{title}{A corpus of {G}erman political speeches from the 21st
  century},
\newblock in: \bibinfo{booktitle}{Proceedings of the Eleventh International
  Conference on Language Resources and Evaluation ({LREC} 2018)},
  \bibinfo{publisher}{European Language Resources Association (ELRA)},
  \bibinfo{address}{Miyazaki, Japan}, \bibinfo{year}{2018}. \URLprefix
  \url{https://aclanthology.org/L18-1127}.
\bibitem[{Mohammad et~al.(2016)Mohammad, Kiritchenko, Sobhani, Zhu, and
  Cherry}]{mohammad-etal-2016-semeval}
\bibinfo{author}{S.~Mohammad}, \bibinfo{author}{S.~Kiritchenko},
  \bibinfo{author}{P.~Sobhani}, \bibinfo{author}{X.~Zhu},
  \bibinfo{author}{C.~Cherry},
\newblock \bibinfo{title}{{S}em{E}val-2016 task 6: Detecting stance in tweets},
\newblock in: \bibinfo{booktitle}{Proceedings of the 10th International
  Workshop on Semantic Evaluation ({S}em{E}val-2016)},
  \bibinfo{publisher}{Association for Computational Linguistics},
  \bibinfo{address}{San Diego, California}, \bibinfo{year}{2016}, pp.
  \bibinfo{pages}{31--41}. \URLprefix \url{https://aclanthology.org/S16-1003}.
  \DOIprefix\doi{10.18653/v1/S16-1003}.
\bibitem[{Kawintiranon and Singh(2022)}]{kawintiranon-singh-2022-polibertweet}
\bibinfo{author}{K.~Kawintiranon}, \bibinfo{author}{L.~Singh},
\newblock \bibinfo{title}{{P}oli{BERT}weet: A pre-trained language model for
  analyzing political content on {T}witter},
\newblock in: \bibinfo{booktitle}{Proceedings of the Thirteenth Language
  Resources and Evaluation Conference}, \bibinfo{publisher}{European Language
  Resources Association}, \bibinfo{address}{Marseille, France},
  \bibinfo{year}{2022}, pp. \bibinfo{pages}{7360--7367}. \URLprefix
  \url{https://aclanthology.org/2022.lrec-1.801}.
\bibitem[{Chen et~al.(2020)Chen, Al~Khatib, Wachsmuth, and
  Stein}]{chen-etal-2020-analyzing}
\bibinfo{author}{W.-F. Chen}, \bibinfo{author}{K.~Al~Khatib},
  \bibinfo{author}{H.~Wachsmuth}, \bibinfo{author}{B.~Stein},
\newblock \bibinfo{title}{Analyzing political bias and unfairness in news
  articles at different levels of granularity},
\newblock in: \bibinfo{booktitle}{Proceedings of the Fourth Workshop on Natural
  Language Processing and Computational Social Science},
  \bibinfo{publisher}{Association for Computational Linguistics},
  \bibinfo{address}{Online}, \bibinfo{year}{2020}, pp.
  \bibinfo{pages}{149--154}. \URLprefix
  \url{https://aclanthology.org/2020.nlpcss-1.16}.
  \DOIprefix\doi{10.18653/v1/2020.nlpcss-1.16}.
\bibitem[{Lai et~al.(2020)Lai, Cignarella, {Hernández Farías}, Bosco, Patti,
  and Rosso}]{lai-muli-stances}
\bibinfo{author}{M.~Lai}, \bibinfo{author}{A.~T. Cignarella},
  \bibinfo{author}{D.~I. {Hernández Farías}}, \bibinfo{author}{C.~Bosco},
  \bibinfo{author}{V.~Patti}, \bibinfo{author}{P.~Rosso},
\newblock \bibinfo{title}{Multilingual stance detection in social media
  political debates},
\newblock \bibinfo{journal}{Computer Speech \& Language} \bibinfo{volume}{63}
  (\bibinfo{year}{2020}) \bibinfo{pages}{101075}. \URLprefix
  \url{https://www.sciencedirect.com/science/article/pii/S0885230820300085}.
  \DOIprefix\doi{https://doi.org/10.1016/j.csl.2020.101075}.
\bibitem[{Barriere et~al.(2022)Barriere, Balahur, and
  Ravenet}]{barriere-etal-2022-debating}
\bibinfo{author}{V.~Barriere}, \bibinfo{author}{A.~Balahur},
  \bibinfo{author}{B.~Ravenet},
\newblock \bibinfo{title}{Debating {E}urope: A multilingual multi-target stance
  classification dataset of online debates},
\newblock in: \bibinfo{booktitle}{Proceedings of the LREC 2022 workshop on
  Natural Language Processing for Political Sciences},
  \bibinfo{publisher}{European Language Resources Association},
  \bibinfo{address}{Marseille, France}, \bibinfo{year}{2022}, pp.
  \bibinfo{pages}{16--21}. \URLprefix
  \url{https://aclanthology.org/2022.politicalnlp-1.3}.
\bibitem[{Bergam et~al.(2022)Bergam, Allaway, and
  Mckeown}]{bergam-etal-2022-legal}
\bibinfo{author}{N.~Bergam}, \bibinfo{author}{E.~Allaway},
  \bibinfo{author}{K.~Mckeown},
\newblock \bibinfo{title}{Legal and political stance detection of {SCOTUS}
  language},
\newblock in: \bibinfo{booktitle}{Proceedings of the Natural Legal Language
  Processing Workshop 2022}, \bibinfo{publisher}{Association for Computational
  Linguistics}, \bibinfo{address}{Abu Dhabi, United Arab Emirates (Hybrid)},
  \bibinfo{year}{2022}, pp. \bibinfo{pages}{265--275}. \URLprefix
  \url{https://aclanthology.org/2022.nllp-1.25}.
\bibitem[{Bundeskanzlei(2018)}]{bk-comm}
\bibinfo{author}{Bundeskanzlei}, \bibinfo{title}{{Die Abstimmungserläuterungen
  des Bundesrates erscheinen im neuen Kleid}}, \bibinfo{year}{2018}. \URLprefix
  \url{https://www.bk.admin.ch/bk/de/home/dokumentation/medienmitteilungen.msg-id-71707.html},
  \bibinfo{note}{accessed: 2022-10-09}.
\bibitem[{Joulin et~al.(2016)Joulin, Grave, Bojanowski, and
  Mikolov}]{joulin2016bag}
\bibinfo{author}{A.~Joulin}, \bibinfo{author}{E.~Grave},
  \bibinfo{author}{P.~Bojanowski}, \bibinfo{author}{T.~Mikolov},
\newblock \bibinfo{title}{Bag of tricks for efficient text classification},
\newblock \bibinfo{journal}{arXiv preprint arXiv:1607.01759}
  (\bibinfo{year}{2016}).
\bibitem[{Pedregosa et~al.(2011)Pedregosa, Varoquaux, Gramfort, Michel,
  Thirion, Grisel, Blondel, Prettenhofer, Weiss, Dubourg, Vanderplas, Passos,
  Cournapeau, Brucher, Perrot, and Duchesnay}]{scikit-learn}
\bibinfo{author}{F.~Pedregosa}, \bibinfo{author}{G.~Varoquaux},
  \bibinfo{author}{A.~Gramfort}, \bibinfo{author}{V.~Michel},
  \bibinfo{author}{B.~Thirion}, \bibinfo{author}{O.~Grisel},
  \bibinfo{author}{M.~Blondel}, \bibinfo{author}{P.~Prettenhofer},
  \bibinfo{author}{R.~Weiss}, \bibinfo{author}{V.~Dubourg},
  \bibinfo{author}{J.~Vanderplas}, \bibinfo{author}{A.~Passos},
  \bibinfo{author}{D.~Cournapeau}, \bibinfo{author}{M.~Brucher},
  \bibinfo{author}{M.~Perrot}, \bibinfo{author}{E.~Duchesnay},
\newblock \bibinfo{title}{Scikit-learn: Machine learning in {P}ython},
\newblock \bibinfo{journal}{Journal of Machine Learning Research}
  \bibinfo{volume}{12} (\bibinfo{year}{2011}) \bibinfo{pages}{2825--2830}.
\bibitem[{Yu(2023)}]{yu-2023-towards}
\bibinfo{author}{Q.~Yu},
\newblock \bibinfo{title}{Towards a more in-depth detection of political
  framing},
\newblock in: \bibinfo{booktitle}{Proceedings of the 7th Joint SIGHUM Workshop
  on Computational Linguistics for Cultural Heritage, Social Sciences,
  Humanities and Literature}, \bibinfo{publisher}{Association for Computational
  Linguistics}, \bibinfo{address}{Dubrovnik, Croatia}, \bibinfo{year}{2023},
  pp. \bibinfo{pages}{162--174}. \URLprefix
  \url{https://aclanthology.org/2023.latechclfl-1.18}.

\end{thebibliography}



\appendix

\onecolumn

\section{Additional label distributions for stance detection for individual languages}
\label{app:boxplots-languages}

\begin{figure*}[h!]
\centering
\includegraphics[width=\linewidth]{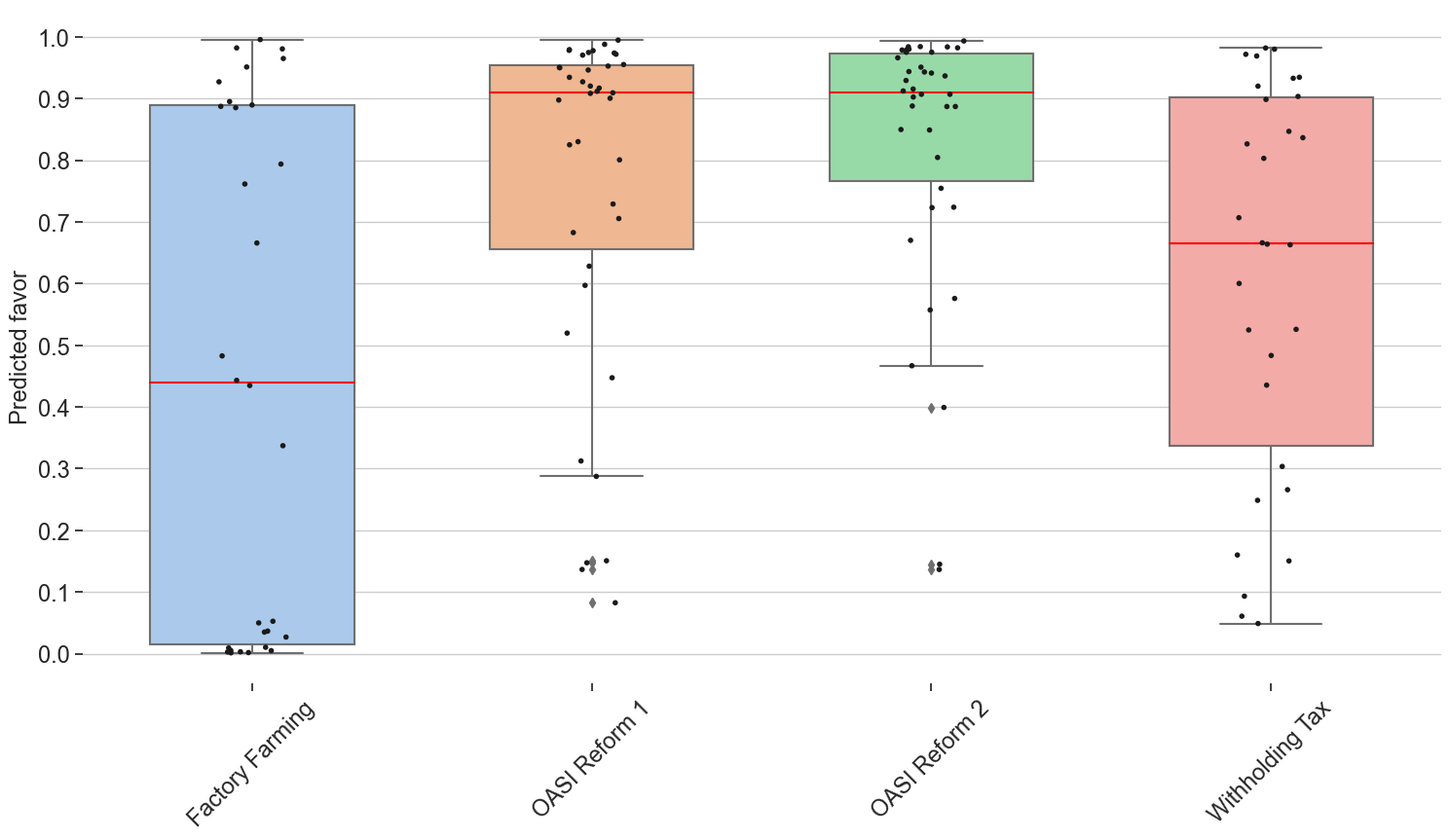}
\caption{Distribution of stances per topic in the \textbf{German} voting booklet. Y values closer to 1 indicate favor. The whiskers extend to the interquartile range. The red line marks the median.}
\label{fig:stance-distribution-de}
\end{figure*}

\begin{figure*}
\centering
\includegraphics[width=\linewidth]{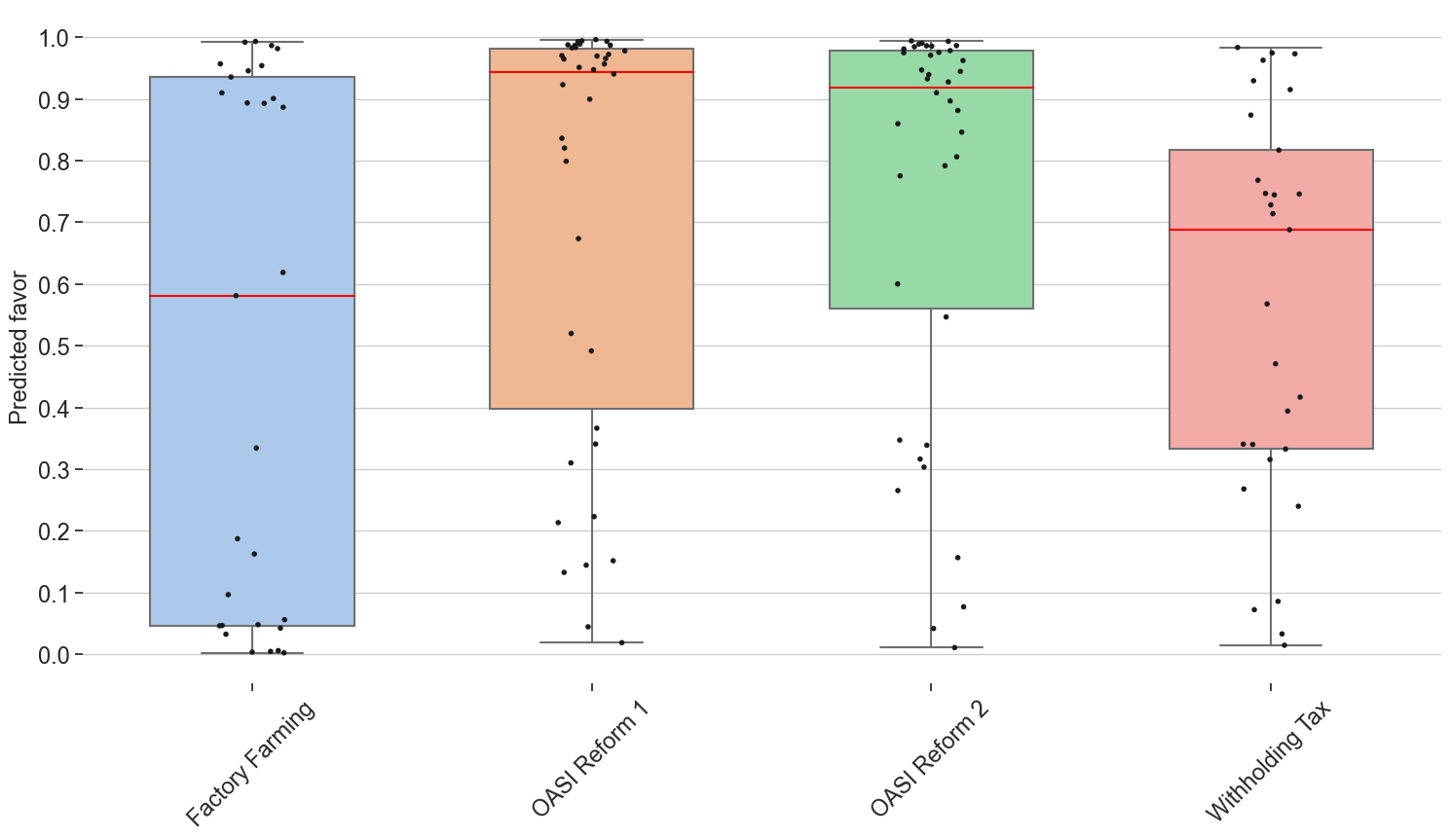}
\caption{Distribution of stances per topic in the \textbf{French} voting booklet. Y values closer to 1 indicate favor. The whiskers extend to the interquartile range. The red line marks the median.}
\label{fig:stance-distribution-fr}
\end{figure*}

\begin{figure*}
\centering
\includegraphics[width=\linewidth]{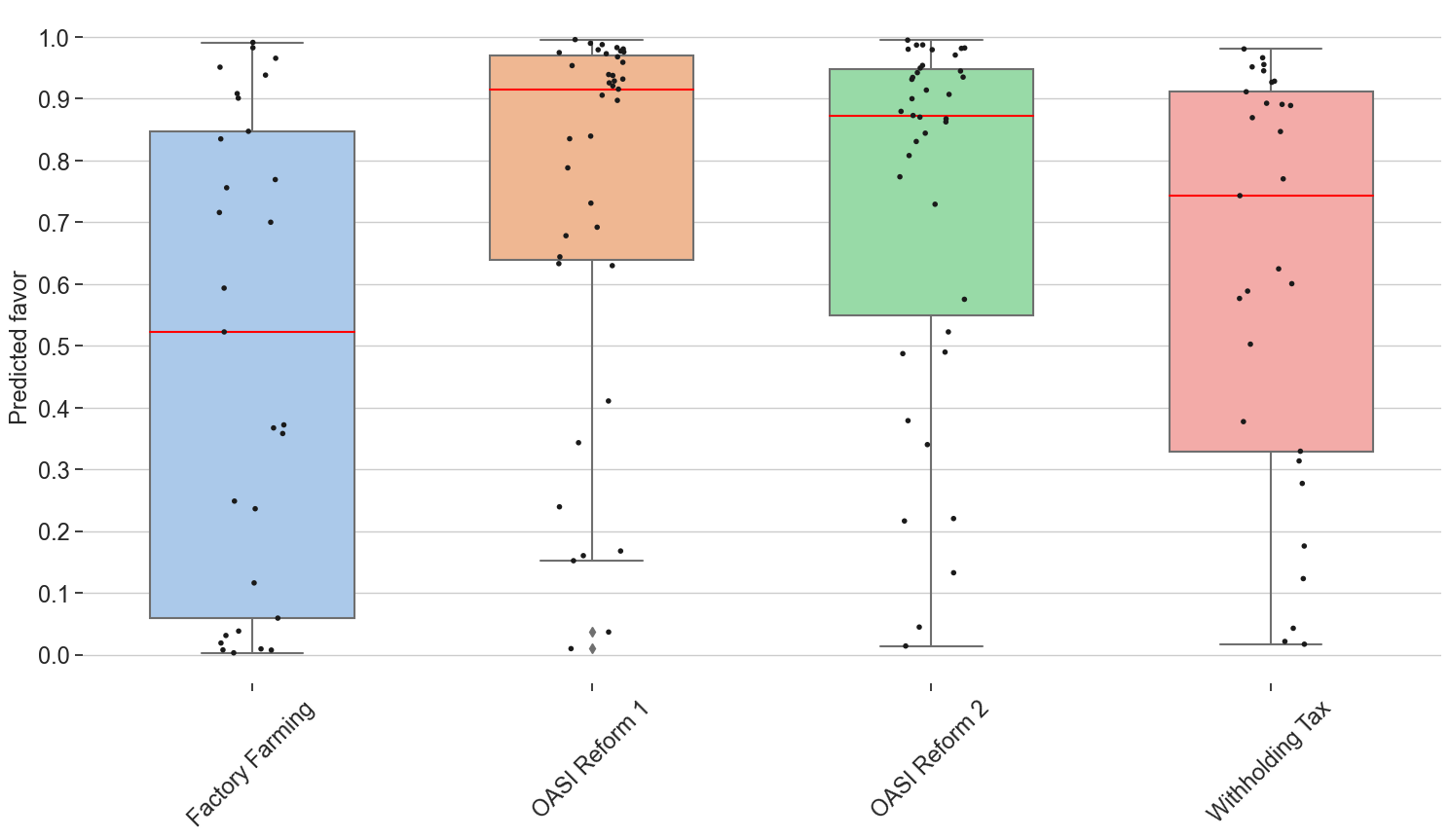}
\caption{Distribution of stances per topic in the \textbf{Italian} voting booklet. Y values closer to 1 indicate favor. The whiskers extend to the interquartile range. The red line marks the median.}
\label{fig:stance-distribution-it}
\end{figure*}

Figures~\ref{fig:stance-distribution-de},~\ref{fig:stance-distribution-fr} and~\ref{fig:stance-distribution-it} display the distribution of stance predictions per topic. While the overall distribution across languages looks similar, the German version includes slightly more statements in favor of each issue, especially with regards to the two OASI reforms.

\end{document}